\documentclass[10pt,twocolumn,letterpaper]{article}

\usepackage{3dv}
\usepackage{times}
\usepackage{epsfig}
\usepackage{graphicx}
\usepackage{amsmath}
\usepackage{amssymb}
\DeclareMathOperator*{\argmax}{arg\,max}
\DeclareMathOperator*{\argmin}{arg\,min}
\usepackage[tight,footnotesize]{subfigure}
\threedvfinalcopy % *** Uncomment this line for the final submission

\def\threedvPaperID{****} % *** Enter the 3DV Paper ID here

\ifthreedvfinal\fi
\begin{document}

%%%%%%%%% TITLE
\title{Unsupervised Temporal Segmentation of Repetitive Human Actions Based on Kinematic Modeling and Frequency Analysis}

\author{Qifei Wang \& Gregorij Kurillo\\
University of California, Berkeley\\
Berkeley, CA 94720, USA\\
{\tt\small \{qifei.wang, gregorij\}@eecs.berkeley.edu}
% To save space, use either the email address or home page, not 
\and
Ferda Ofli\\
Qatar Computing Research Institute\\
Doha, Qatar\\
{\tt\small fofli@qf.org.qa}
\and
Ruzena Bajcsy\\
University of California, Berkeley\\
Berkeley, CA 94720, USA\\
{\tt\small bajcsy@eecs.berkeley.edu}
\thanks{This research was supported by the National Science Foundation (NSF) under Grant No. 1111965.}
}

\maketitle
%\thispagestyle{empty}

%%%%%%%%% ABSTRACT
\begin{abstract}
   In this paper, we propose a method for temporal segmentation of human repetitive actions based on frequency analysis of kinematic parameters, zero-velocity crossing detection, and adaptive k-means clustering. Since the human motion data may be captured with different modalities which have different temporal sampling rate and accuracy (e.g., optical motion capture systems vs. Microsoft Kinect), we first apply a generic full-body kinematic model with an unscented Kalman filter to convert the motion data into a unified representation that is robust to noise. Furthermore, we extract the most representative kinematic parameters via the primary frequency analysis. The sequences are segmented based on zero-velocity crossing of the selected parameters followed by an adaptive k-means clustering to identify the repetition segments. Experimental results demonstrate that for the motion data captured by both the motion capture system and the Microsoft Kinect, our proposed algorithm obtains robust segmentation of repetitive action sequences.
\end{abstract}

%%%%%%%%% BODY TEXT
\section{Introduction}

Temporal motion segmentation is a crucial step in human motion understanding and analysis \cite{moeslund2006survey}. In many applications of human-machine interaction, such as gesture-based computer interaction, robotic manipulation, and computer gaming, the humans have to perform a set of primitive actions multiple times. This is in particular the case in various interactive fitness or rehabilitation applications where participants are required to perform several repetitive exercises while their physical performance is being monitored by motion sensors. To provide feedback on the performance (e.g., automatic repetition counting) or to perform post-exercise performance analysis, it is necessary to partition the repetitive motion data into multiple segments where each represents one temporal repetition of the primitive action.

Extensive literatures have addressed the problems of temporal motion segmentation. Jenkins et al. \cite{fod2002automated} proposed a zero-velocity crossing (ZVC) detection algorithm based on joint angle velocities to partition the motion data of repetitive arm exercises into individual repetitions. Due to high sensitivity of the zero-velocity crossing to input noise, it is only practical when direct velocity measurements are available or when the position measurements have small jitter. Later on, majority of temporal segmentation efforts have been focused on the motion data captured by an optical motion capture system since it can provide motion data with high frame rate ($>100Hz$) and high positional accuracy (up to $\sim1mm$) \cite{dyer1995motion}. Lu and Ferrier \cite{lu2004repetitive} introduced a multi-dimensional segmentation algorithm to automatically decompose a complex motion into a sequence of simple linear dynamical models. Additionally, Jernej \textit{et al}. \cite{barbivc2004segmenting} proposed three segmentation methods based on the principle component analysis (PCA) to distinguish one primitive action from the other. Their first two methods can perform the segmentation in real-time using PCA and probabilistic PCA, respectively, whereas the third method (PCA-GMM) fits a Gaussian mixed model to the data of the entire exercise sequence offline. Although the proposed PCA-GMM segmentation outperforms the previous two, the PCA-based approach cannot always distinguish the segments of repetitive actions since the principle components within each segment are typically not sufficiently distinguishable. Zhou et al. \cite{zhou2013hierarchical} proposed a bottom-up hierarchical aligned clustering analysis (HACA) algorithm by combining kernel $k$-means with generalized dynamic time alignment kernel to cluster motion data into motion primitives. Other unsupervised methods intended for temporal segmentation of activities into distinct actions include neighbor graph \cite{vogele2014efficient}, ZVC with hidden Markov model \cite{lin2014online}, and continuous linear dynamic system \cite{wang2013human}. These methods, however, typically cannot segment the periodic actions into repetitive primitives because of large similarity between the action units.

More recently, depth-sensing cameras such as Microsoft Kinect \cite{zhang2012microsoft} have been introduced as a convenient low-cost alternative to full-scale motion capture systems for many real-world applications. This type of sensor captures both the texture images and the depth information from the observed scene, and extracts the human pose \cite{shotton2013real} in real time. Since the data captured by Kinect are noisy (with joint position errors of several centimeters) and have relatively low framerate (30 FPS) as compared to the marker-based motion capture, the motion segmentation is even more challenging. Several researchers approached the human action recognition and motion analysis of Kinect data with supervised methods, e.g. human action recognition based a two-layer hierarchical hidden Markov model (HMM) \cite{sung2012unstructured}, and gesture recognition based on a cascaded correlation-based classifier \cite{raptis2011real}. The supervised approaches, however, are time consuming and require considerable amount of training data which may be difficult to match the performance differences across users. Furthermore, these literatures do not consider the repetitive actions.

As majority of the segmentation frameworks address the partitioning of motion sequences into distinct actions or action primitives, less attention has been given to the segmentation of repetitive actions into individual repetitions. In this paper, we propose a generic unsupervised segmentation approach based on the inherent properties of such repetitive motion. Our motion segmentation algorithm can be summarized as follows:
a)	Application of a generic kinematic model using unscented Kalman filter (UKF) \cite{wan2000unscented} to convert the motion data into a unified representation that also reduces the effects of noise;
b)	Use frequency analysis of repetitive motion data to determine the most representative kinematic parameters for repetition segmentation;
c)	Finally, application of zero-velocity crossing detection followed by an adaptive k-means clustering to obtain robust repetition segmentation in accordance with the motion phase.

The main contributions of this work are: (1) robust segmentation of repetitive actions under large input data noise; (2) no need for training data or manual annotation; (3) general approach for both optical motion capture and Kinect type of modalities.

The rest of this paper is organized as follows: Section 2 briefly describes the proposed segmentation framework. The kinematic model and the unified data transformation based on UKF are introduced in Section 3. Section 4 provides details on the temporal segmentation based on the frequency analysis, zero-velocity crossing detection and adaptive $k$-means clustering. Finally, Section 5 demonstrates the experimental results on motion capture and Kinect datasets and Section 6 concludes the paper.

\section{Temporal repetitive action segmentation framework}

In this section, we provide an overview of the proposed framework for temporal segmentation of repetitive human motion. Each proposed module will be further described in subsequent sections.

\begin{figure}[t]
\begin{center}
   \includegraphics[width=0.9\linewidth]{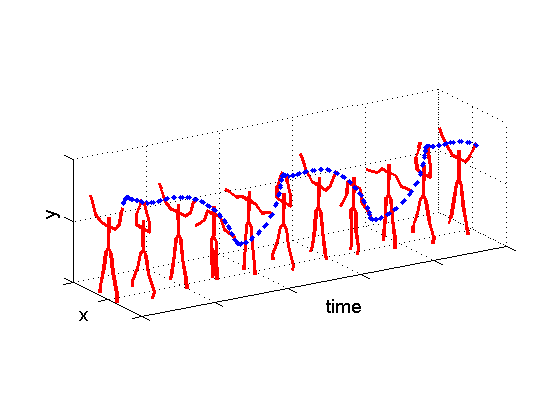}
\end{center}
   \caption{An example of a repetitive motion sequence ``\textit{jumping jacks}'' captured by an optical motion capture system. The dotted curve demonstrates the trajectory of the left hand.}
\label{fig1}
\end{figure}

In this paper, we address the input motion data represented as a sequence of skeletal poses. Figure \ref{fig1} demonstrates an articulated skeleton sequence of the repetitive action, ``\textit{jumping jacks}'', based on the joint trajectories.

\begin{figure}[t]
\begin{center}
\includegraphics[width=0.9\linewidth]{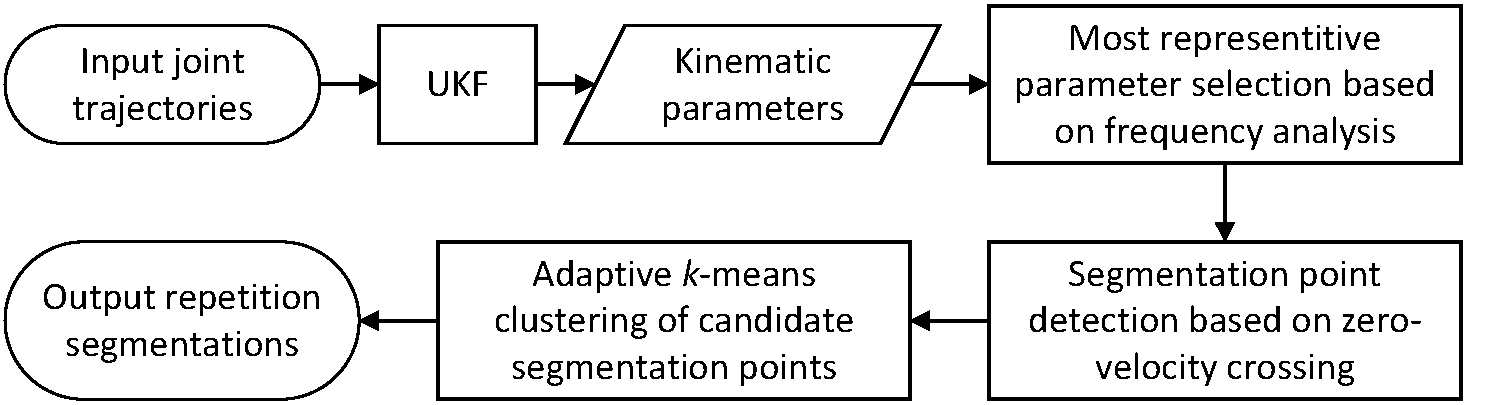}
\end{center}
\caption{Overview of the temporal repetitive action segmentation framework.}
\label{fig2}
\end{figure}

In the first step of our segmentation framework, the input joint trajectories are converted into the parameters of a unified kinematic model by a four-pass UKF. After the UKF processing, we select the most representative kinematic parameters that best capture the motion repetitions based on the frequency analysis. The zero-velocity crossing detection is used to identify possible segmentation candidates in each of the selected kinematic parameter sequences. During this step, multiple candidate points for segmentation may be generated in various stages of the motion since the human typically pauses briefly while transitioning between different phases of motion. To consolidate the segmentation points, we further apply an adaptive $k$-means clustering algorithm to determine the boundaries of each repetition. The framework of the proposed method is summarized in Figure \ref{fig2}.

\section{Kinematic filtering with UKF}

Since the human motion data may be captured by different modalities which may have different skeletal configurations, temporal sampling rates, and accuracy, transforming the captured data to a unified kinematic representation can alleviate the differences between the motion capture modalities. Such representation can facilitate the development of a generic motion segmentation approach. In this paper, we extend the kinematic model proposed in \cite{matthew2014calculating} from upper extremity to a full-body kinematic model. The kinematic filtering with UKF is intended to transform the motion data of joint positions into the kinematic parameters while suppressing noise. In this section, we briefly introduce the kinematic model and demonstrate its performance on the data captured by the marker-based motion capture system and Kinect.

\subsection{Kinematic model}

In motion analysis, the human body can be represented by a series of bones connected via joints. We can thus create a kinematic chain to model the motion of the limbs relative to the selected root body segment (e.g. torso). The location of each joint in the chain can be derived by its parent joint position, rotation, and length of the bone segment which connects the current joint to its parent. In this paper, we model the upper extremity using a 6-DoF kinematic model and the lower extremity by a 4-DoF kinematic model. Other more or less complex models can also be used since our segmentation approach is independent of the selected kinematic representation. Figure 3 shows the kinematic model used in our analysis on the left upper and lower extremities. That model is the same for the right part. For the upper extremity, the length of clavicle, humerus, and radius are denoted by $l_c$, $l_h$, $l_r$, respectively. The shoulder is modeled as a spherical joint, with three DoFs, which are denoted by a triplet of angles, $\textbf{r}_{sho}=(r_{sho}^X,r_{sho}^Y,r_{sho}^Z)$, representing the rotation angles about each axis. Two DoFs denoted by $\textbf{r}_{sca}=(r_{sca}^X,r_{sca}^Y)$ are used to represent the flexion/extension and abduction/adduction angles of the scapula. Finally, a single DoF model is used to represent the flexion/extension angle, $r_{elb}$, of the elbow joint.

Given the scapular position from the world origin, denoted by $\textbf{p}_{sca}=(p_{sca}^X,p_{sca}^Y,p_{sca}^Z)$, and the joint positions of shoulder ($\textbf{p}_{sho}$), elbow ($\textbf{p}_{elb}$), and wrist ($\textbf{p}_{wri}$), are represented by the following functions, respectively:

\begin{equation}
\label{eq1}
\textbf{p}_{sho}=H_1(\textbf{p}_{sca},l_c,\textbf{r}_{sca}),
\end{equation}
\begin{equation}
\label{eq2}
\textbf{p}_{elb}=H_2(\textbf{p}_{sca},l_c,l_h,\textbf{r}_{sca},\textbf{r}_{sho}),
\end{equation}
\begin{equation}
\label{eq3}
\textbf{p}_{wri}=H_3(\textbf{p}_{sca},l_c,l_h,l_r,\textbf{r}_{sca}  ,\textbf{r}_{sho},r_{elb}).
\end{equation}
In (\ref{eq1}), (\ref{eq2}) and (\ref{eq3}), $H_1$, $H_2$, and $H_3$ are the kinematic transformation functions for shoulder, elbow, and wrist, respectively.

\begin{figure}[t]
\begin{center}
\includegraphics[width=0.9\linewidth]{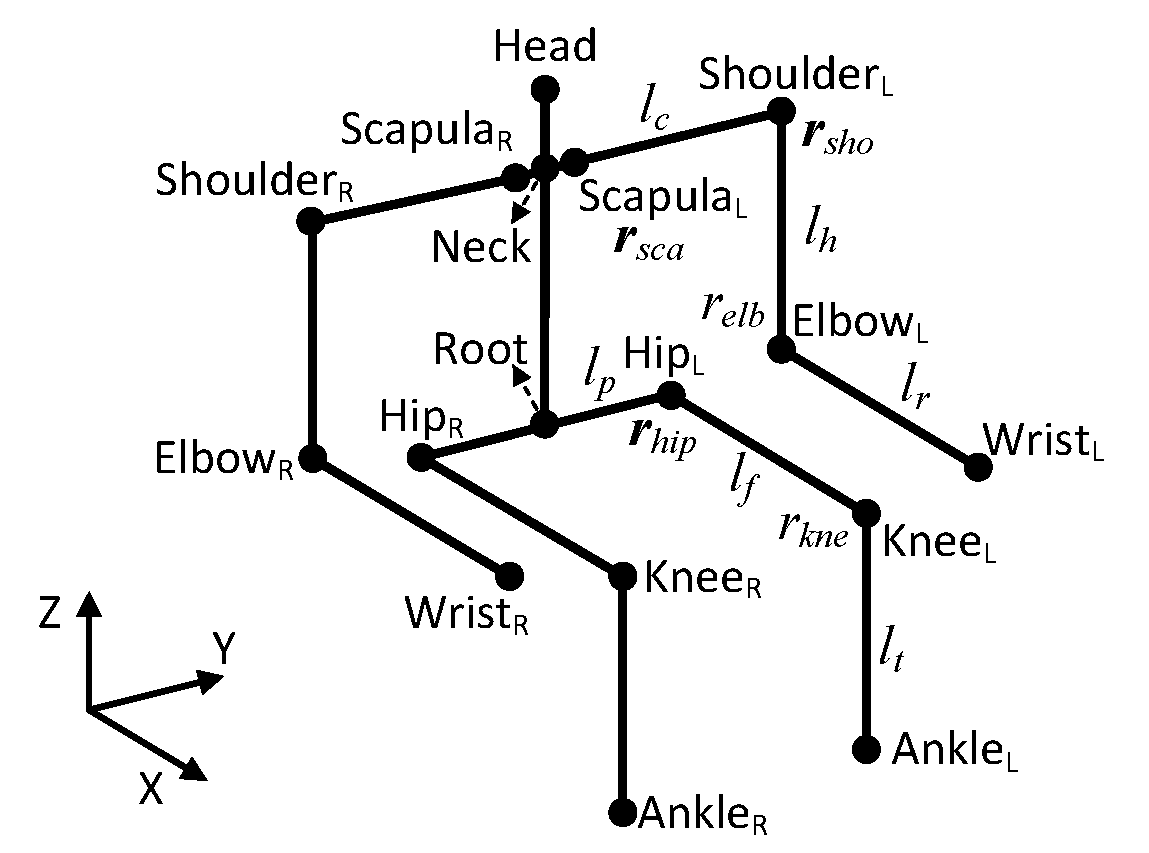}
\end{center}
\caption{Overview of the temporal repetitive action segmentation Kinematic model of human skeletal pose. The subscripts, ``L'' and ``R'', denote the left and right, respectively. The kinematic parameters are only labeled on the left upper and lower extremities. The right upper and lower extremities share the same kinematic models as the left ones.}
\label{fig3}
\end{figure}

As shown in Figure \ref{fig3}, the kinematic model of the lower extremity is represented as follows: the length of pelvic, femur, and tibia are denoted by $l_p$, $l_f$, and $l_t$, respectively. The hip is modeled as a 3-DoF joints by $\textbf{r}_{hip}=(r_{hip}^X,r_{hip}^Y,r_{hip}^Z )$ to represent the rotation about the three axes. A single DoF model is used to represent the flexion/ extension angle, $r_kne$, of the knee joint. Given the hip joint position, $\textbf{p}_{hip}$, the positions of knee ($\textbf{p}_{kne}$) and ankle ($\textbf{p}_{ank}$) can be derived by the two kinematic functions, $H_4$ and $H_5$, as follows:

\begin{equation}
\label{eq4}
\textbf{p}_{kne}=H_4(\textbf{p}_{hip},l_p,l_f,\textbf{r}_{hip}),
\end{equation}
\begin{equation}
\label{eq5}
\textbf{p}_{ank}=H_5(\textbf{p}_{hip},l_p,l_f,l_t,\textbf{r}_{hip},r_{kne}).
\end{equation}

\subsection{Kinematic parameter estimation}

In this paper, we apply a four-pass UKF to estimate the kinematic parameters based on the input joint trajectories. In the first two passes, the UKF is performed forward and backward respectively. The state vector at time $t$, $\textbf{x}$(t), is composed of all the kinematic parameters of the upper and lower extremities. The observation vector, $\textbf{y}$(t), is concatenated by the coordinate vectors of the joints of the upper and lower extremities. In order to adapt the kinematic model to any type of motion, the state transition of each parameter is modeled by a random-walk process. Therefore, the state prediction and observation measuring functions of the forward UKF can be represented as:

\begin{equation}
\label{eq6}
\textbf{x}(t)=\textbf{x}(t-1)+\boldsymbol{\eta}(t),
\end{equation}
\begin{equation}
\label{eq7}
\textbf{y}(t)=\textbf{\textit{H}}(\textbf{x}(t))+\boldsymbol{\xi}(t).
\end{equation}

In (\ref{eq6}) and (\ref{eq7}), $\boldsymbol{\eta}(t)$ and $\boldsymbol{\xi}(t)$ denote the noise terms of state transition and observation measurement, respectively. The function \textbf{\textit{H}} is the combination of the kinematic transform functions $H_1$  , $H_2$, $H_3$, $H_4$, and $H_5$. In the backward filtering, the state prediction function is
\begin{equation}
\label{eq8}
\textbf{x}(t)=\textbf{x}(t+1)+\boldsymbol{\eta'}(t),
\end{equation}
$\boldsymbol{\eta'}(t)$ as above denotes the noise term of the measuring function. The observation function is kept the same as in (\ref{eq7}).

In the rigid-body kinematic model, the bone length of each body segment should not change during movement. Due to the noise of the motion data, especially in Kinect, the bone lengths of the input data may vary considerably. Since it is not practical to obtain the accurate bone length of each person in advance, we thus perform the two more passes of the filtering with UKF. In the second forward and backward passes, the bone lengths, $l_c$, $l_h$, $l_r$, $l_p$, $l_f$, and $l_t$ are fixed to the optimized values which are the expectation of the estimated bone lengths obtained in the first two passes. The corresponding state prediction parameters of $l_c$, $l_h$, $l_r$, $l_p$, $l_f$, and $l_t$ in $\boldsymbol{\eta}(t)$ and $\boldsymbol{\eta}(t)$ are set to zero during the last two passes.

\begin{figure}[!htbp] 
	\begin{tabular}{cc}
		\includegraphics[width=0.47\linewidth]{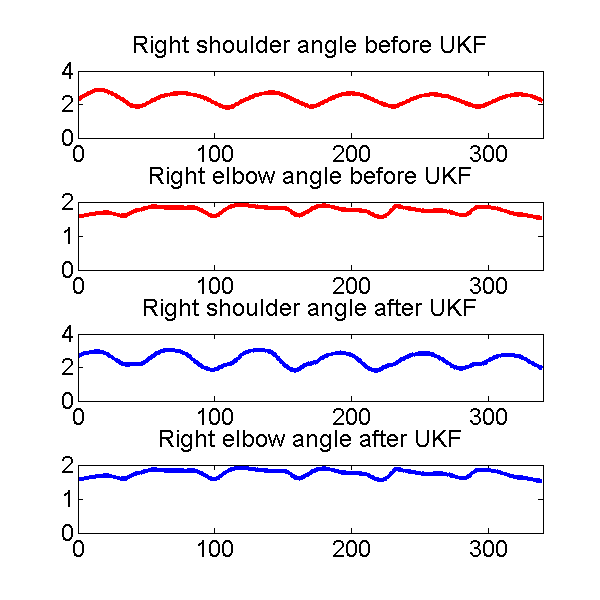}&
		\includegraphics[width=0.47\linewidth]{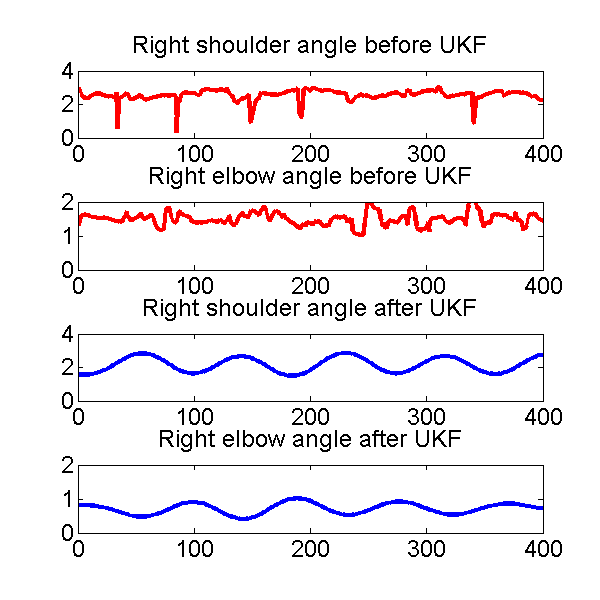}\\
		(a)  & (b)   \\
	\end{tabular}
	\caption{Joint angles of the shoulder and elbow calculated by the input motion data and the output kinematic parameters, respectively, horizontal axes represent the frame count, vertical axes represent the joint angle, (a) sequences captured by motion capture system, (b) sequences captured by Kinect.}
	\label{fig4}
\end{figure}

Figure \ref{fig4} compares the joint angles derived from the input joint trajectories and the output kinematic parameters. For the data from the motion capture system (Figure \ref{fig4}a), the joint angles derived from the input joint positions clearly indicate periodic nature of the movement. After the transformation, the joint angle curves maintain similar periodic pattern as the input data. For the motion data captured by Kinect (Figure \ref{fig4}b), the joint angles derived from the input joint positions are much noisier, making it more difficult to detect periodic characteristics as compared to the input data from the motion capture system. After applying the proposed model, the periodic patterns are much more pronounced. This example clearly shows that the kinematic model can convert the motion data captured with different levels of accuracy into a unified representation that preserves the periodicity and smoothness of the motion.

\section{Repetition segmentation based on kinematic modeltitle}

In this section, we provide a detailed description of the three steps for the repetition segmentation from the given kinematic parameters.

\subsection{Most representative kinematic parameter selection}

For a certain action, not all DoFs will be active. The motion segmentation can thus be defined only by certain parameters which exhibit periodic behavior. We refer to these parameters as the \textit{most representative kinematic parameters}.

In order to select the most representative set of parameters, we propose a frequency domain ranking algorithm. For each parameter, $x_i$, we perform Fourier transform to the temporal data, $f_i(t)$. Next, we normalize the amplitude and obtain the frequency response, $\hat{f}_i(\omega)$. Here, $t$ and $\omega$ denote the time stamp and frequency, respectively. We sum the power of all the kinematic parameters with respect to every discrete frequency point and determine the nonzero frequency with the maximum sum of the power. That nonzero frequency with the maximum power is called the \textit{primary frequency}, $\omega_p$, which is also used to approximate the frequency of the primitive action within the exercise sequence. The primary frequency can be obtained as follows:

\begin{equation}
\label{eq9}
\omega_p=\argmax_{\omega}\sum_i\|\hat{f}_i(\omega)\|^2.
\end{equation}

\begin{figure}[!htbp] 
	\begin{tabular}{cc}
		\includegraphics[width=0.47\linewidth]{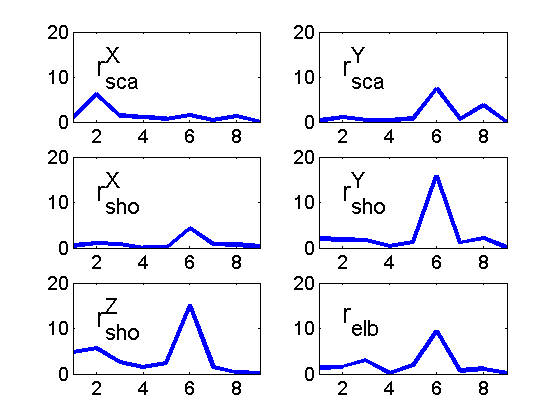}&
		\includegraphics[width=0.47\linewidth]{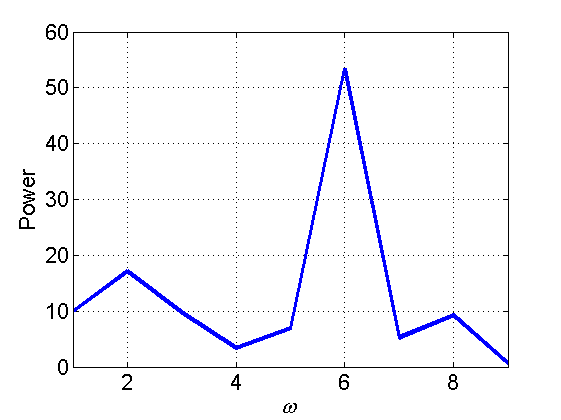}\\
		(a)  & (b)   \\
	\end{tabular}
	\caption{Frequency domain analysis of the kinematic parameters: (a) Fourier transform of the kinematic parameters of right arm; (b) power sum of all the kinematic parameters.}
	\label{fig5}
\end{figure}

Figure \ref{fig5} demonstrates an example for the primary frequency detection. The curves in Figure \ref{fig5}a show the frequency response amplitudes of the six parameters of the upper extremity for a repetitive motion sequence. Figure \ref{fig5}b demonstrates the sum of the power for all the parameters with respect to the frequency. We can observe that the power at the frequency six is the largest one for the whole sequence. Therefore, the primary frequency for this sequence is set to six. Subsequently, we sort all the parameters according to their power of the primary frequency. In order to obtain robust repetition segmentation, we need to select multiple parameters for segmentations rather than the one with the largest power at the primary frequency. The number of selected parameters is determined by the following criterion. For the primary frequency, when the ratio between the accumulated power of the top $M$ parameters and the power sum of all the parameters is larger than the power ratio threshold $\theta$, we choose the top $M$ parameters as the input to the following segmentation steps.

\subsection{Segmentation point detection}

During a repetitive action, the movement of the limbs/joints often changes direction or pauses when transitioning from one cycle to the other. According to this observation, we develop a segmentation point detection algorithm based on the zero-velocity crossing detection.

Due to the noise in the output parameter sequences of UKF, especially for the noisy data captured by Kinect, it is necessary to perform a band-pass filtering on each sequence of the selected parameters to remove remaining jitter which may adversely affect the performance of the zero-velocity crossings detection. For the analysis shown in this paper, we used a band-pass Butterworth filter. The center of the passing band window is set to the primary frequency obtained in the previous step, and the window width is empirically set to 3. The window width represents a tolerance to the variability of the repetition segment lengths in a sequence.

After performing the band-pass filtering, we calculate the first-order derivative $\{\Delta_m(t),m=1,2,\cdots,M\}$ of the sequences of selected parameters, $\{x_m'(t),m=1,2,\cdots,M\}$, representing the corresponding velocity. Ideally, all the velocities should reach zero at the same time. Because of noise, the zero-crossing detection criterion is relaxed to the squared sum of all the velocities reaching the local minimum value which can be represented as

\begin{equation}
\label{eq10}
t_c=\argmin_{t \in [t_1, t_2]}⁡\sum_{m=1}^M\Delta_m^2(t)).
\end{equation}
In (\ref{eq10}), $[t_1,t_2]$ denotes a sliding window with overlap to obtain the time stamp $t_c$ which achieves the local minimal squared sum of velocities. Each $t_c$ is treated as a candidate segmentation point of the motion sequences.

\subsection{Adaptive \textbf{\textit{k}}-means clustering}

Since the motion sequence may contain multiple brief pauses during the transitions between different phases of motion, the zero-velocity crossing detector may detect multiple candidate points for the segmentation. The candidate points will thus result in over-segmentation of the activity sequence. The over-segmentation can be addressed by clustering these points into several groups and partitioning the sequences based on one group. However, as various motion sequences may have different numbers of transitions within one cycle, the number of the clusters is unknown and also difficult to predict before processing the data. Therefore, an adaptive $k$-means algorithm is proposed for the task of candidate segmentation point clustering.

Suppose the number of the candidate segmentation points is $N$. The input data samples to the clustering algorithm are the vectors of the selected parameters for the candidate segmentation points. Denote the vector samples of the candidate segmentation points as $\textbf{v}_n$,$(n=1,2,\cdots,N)$. If the number of the clusters is defined as $K$, all the candidate segmentation points will be classified into $K$ classes by the $k$-means clustering algorithm. Afterwards, the inter-class and intra-class distances of such $k$-means clustering can be defined by the following functions:

\begin{equation}
\label{eq11}
J_{intra}(K)=\sum_{k=1}^{K}\sum_{n=1}^{N}g_{kn}\|\textbf{v}_n-\textbf{u}_k\|^2,
\end{equation}

\begin{equation}
\label{eq12}
J_{inter}(K)=\sum_{k=1}^{K}\sum_{j=1}^{K}\|\textbf{u}_j-\textbf{u}_k\|^2.
\end{equation}

In (\ref{eq11}) and (\ref{eq12}), $\textbf{u}_k$ denotes the center of the $k$-th cluster, the parameters $\{g_{kn}\}$ in (\ref{eq11}) form a binary indicator matrix, $\epsilon\{0,1\}^{K\times N}$, such that $g_{kn}=1$ if the sample $\textbf{v}_n$ belongs to cluster $\textbf{u}_k$ and zero otherwise. Therefore, the intra-class distance is defined as the sum of the Euclidean distance between the $n$-th sample parametric vector and its corresponding cluster center. On the other hand, the inter-class distance is defined as the sum of the Euclidean distance between the centers of any two clusters. As the optimal number of clusters is unknown, the number of clusters, $K^*$, which also represents the numbers of action phases, can be defined as the one that can minimize the overall intra- and inter- class distance cost function as:

\begin{equation}
\label{eq13}
K^*=\argmin_{K}J_{intra}(K)+\lambda\times J_{inter}(K),
\end{equation}
where the parameter $\lambda=N/K^2$ in (\ref{eq13}) is a weighting coefficient. Based on the empirical study, the largest K should not exceed 10, and the distance cost function in (\ref{eq13}) is usually a convex function. Therefore, the optimization problem in (\ref{eq13}) can be solved by an iterative search algorithm which initializes $K=2$ and increases $K$ by one until the distance cost function reaches the minimum value.

After obtaining the cluster number, the algorithm perform $k$-means clustering of all the candidate segmentation points and generate $K^*$ groups of points. The final segmentation points are selected as the points in the class which can span majority of the frames in the sequences. The span of one segmentation point is defined as the number of frames whose distance to that segmentation point is less than $\psi/\omega_p$ , where $\psi$ is the number of frames in that sequence and $\omega_p$ is the primary frequency. The overall span of one group of candidate segmentation points is the number of frames spanned by all the points in that group. Therefore, the final selection criterion is defined as follows:

\begin{equation}
\label{eq14}
k^*=\argmax_{1\leq k\leq K^*}(\phi(t_c^k)).
\end{equation}
In (\ref{eq14}), $\phi(t_c^k)$ denotes the total frame number spanned by the candidate segmentation points in the $k$-th group, $k^*$ denotes the index of the candidate segmentation point group.

\begin{figure}[!htbp] 
	\begin{tabular}{cc}
		\includegraphics[width=3.9cm]{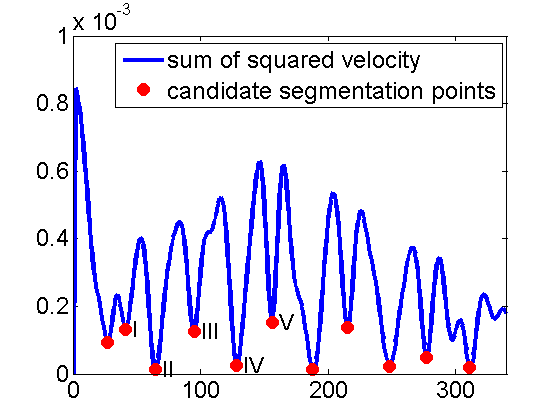}&
		\includegraphics[width=3.9cm]{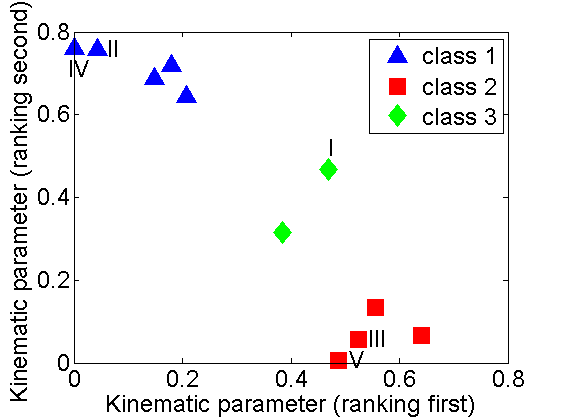}\\
		(a)  & (b)  \\
	\end{tabular}
	
	\begin{tabular}{ccccc}
		\\
		\includegraphics[width=1.3 cm]{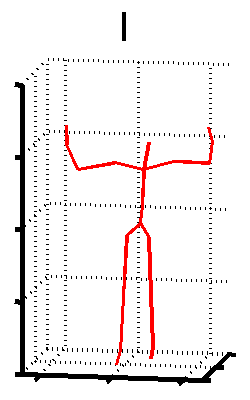}&
		\includegraphics[width=1.3 cm]{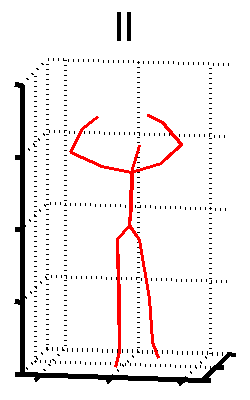}&
		\includegraphics[width=1.3 cm]{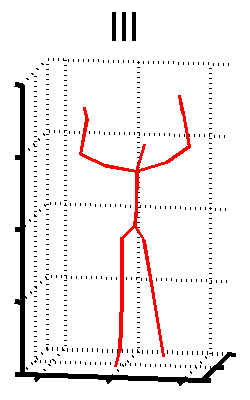}&
		\includegraphics[width=1.3 cm]{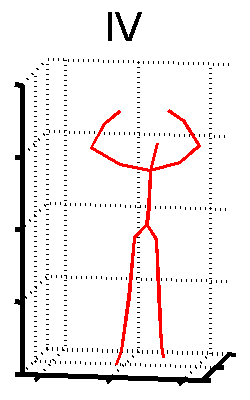}&
		\includegraphics[width=1.3 cm]{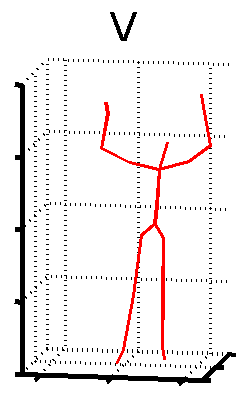}
		\\
		\multicolumn{2}{}{} & (c) & \multicolumn{2}{}{} \\
	\end{tabular}
	\caption{Adaptive $k$-means clustering of candidate segmentation point clustering: (a) candidate segmentation points on the sum of squared velocity curve; (b) clustering of the candidate segmentation points represented as 2D plot; (c) identified skeleton configurations at the corresponding candidate segmentation points.}
	\label{fig6}	
\end{figure}

Figure \ref{fig6} demonstrates the results of the segmentation point clustering on the sequence ``\textit{clapAboveHead5Reps}'' from HDM05 database \cite{muller2007documentation}. The circular points in Figure \ref{fig6}a are the candidate segmentation points detected by the zero-velocity crossing detector. The adaptive $k$-means clustering results are shown in Figure \ref{fig6}b, where each point is corresponding to one of the candidate segmentation points in Figure \ref{fig6}a. All the points are finally classified into three categories by the adaptive $k$-means clustering. Based on the proposed selection criterion, we select points in the class \#2 for the final segmentation. Figure \ref{fig6}c demonstrates the skeleton configurations corresponding to the labeled candidate segmentation points from Figures \ref{fig6}a and \ref{fig6}b. We can observe that the corresponding skeleton configurations in each cluster belong to the same state of activity.

\section{Experimental results}

In this section we present experimental results on the motion sequences captured by both motion capture system and Kinect. We select five sequences from HDM05 \cite{muller2007documentation} database: \textit{clapAboveHead5Reps, clap5Reps, jumpingJack-3Reps, rotateArmsBothBackward3Reps, elbowToKnee-3RepsLelbowStart}. The first two sequences contain 5 repetitions and the last three contain 3 repetitions. Each sequence is performed by 5 subjects. We also collected our own repetitive motion database which contains 10 repetitive exercises: \textit{Shallow Squats, Chair Stands, Buddha’s Prayer, Cops \& Robbers, Abs in Knee Lifts, Lateral Stepping, Clapping, Punching, Line Stepping, and Pendulum}, performed by 10 subjects where each exercise action is repeated 5 times. The motion data were recorded simultaneously by the optical motion capture system and Kinect camera. In the reminder of this paper, if not specified, the sequences from our database refer to the motion data captured by Kinect. To obtain the ground truth segmentation of actions into repetitions, we manually segmented each skeletal sequence by observing the corresponding video data and marking the frames that correspond to the start/end of each repetition segment.

We evaluated three temporal repetition segmentation algorithms, including the algorithm based on PCA and GMM \cite{barbivc2004segmenting}, HACA \cite{zhou2013hierarchical}, and our proposed algorithm. The results of the segmentations are compared with the ground truth, provided by the manual segmentation. For simplicity, the above four methods are denoted by ``PCA-GMM'', ``HACA'', ``Proposed'', and ``Manual'' in the remainder of the paper.

\begin{figure*}[!htbp]
	\centering
	\begin{tabular}{cc}
		\includegraphics[width=8.6cm]{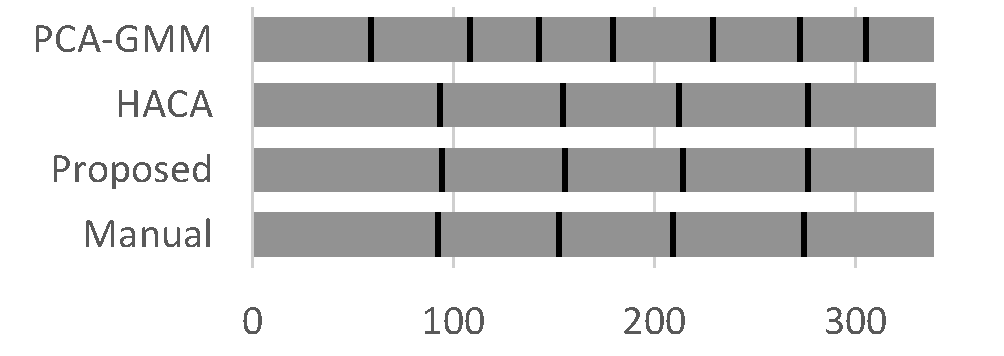}&
		\includegraphics[width=8.6cm]{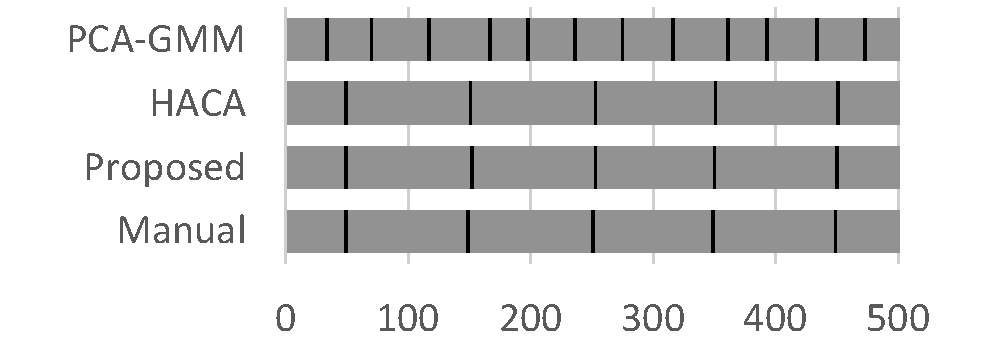}\\
		(a) action: ``\textit{clapAboveHead5Reps}'', actor: ``bd'' & (b) action: ``\textit{clapAboveHead5Reps}'', actor: ``dg'' \\
		\includegraphics[width=8.6cm]{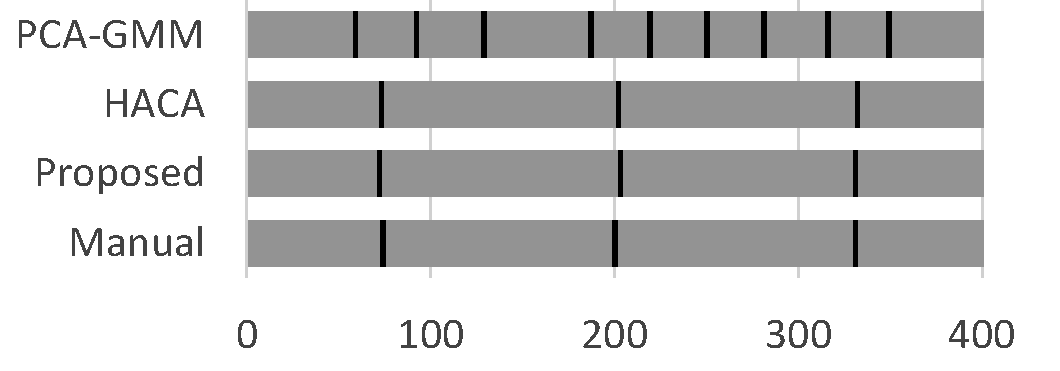}&
		\includegraphics[width=8.6cm]{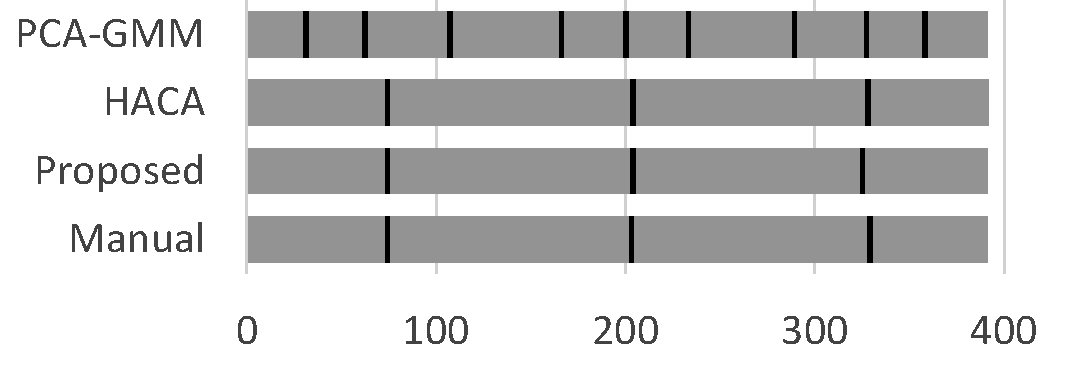}\\
		(c) action: ``\textit{jumpingJack3Reps}'', actor: ``bd''  & (d) action: ``\textit{jumpingJack3Reps}'', actor: ``dg'' \\
		\multicolumn{2}{c}{\includegraphics[width = 6cm]{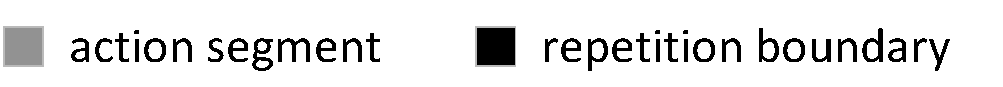}}\\
	\end{tabular}
	\caption{Temporal repetition segmentation results of the four approaches, Manual, Proposed, HACA \cite{zhou2013hierarchical}, and PCA-GMM \cite{barbivc2004segmenting}, on the sequences from HDM05 database which are acquired by motion capture system.}
	\label{fig7}
\end{figure*}

Figure \ref{fig7} demonstrate repetition segmentation results for the sequences from HDM05 database acquired by motion capture system. Figures \ref{fig7}a and \ref{fig7}b show the results for the sequence ``\textit{clapAboveHead5Reps}'' performed by the two actors, ``bd'' and ``dg'', respectively. Figures \ref{fig7}c and \ref{fig7}d show the results for the sequence ``\textit{jumpingJack3Reps}'', also performed by the same actors. For clarity, in each sequence, only the frames with indices smaller than 400 are shown in the figures. From the segmentation results, we can observe that, for the motion capture data with high precision, HACA and our proposed algorithm can obtain approximately the same segmentation as the manual approach. However, PCA-GMM usually over-segments the sequences and has considerable false detection rate of the segmentation points. The over-segmentation is a result of high similarity of the motion data of different repetitions, thus making the features of different repetitions not distinguishable enough in the subspace generated by PCA and GMM. Therefore, this method cannot robustly detect the transition between two repetitions unless the actor performed the same action with quite different style.

\begin{figure*}[!htbp]
	\centering
	\begin{tabular}{cc}
		\includegraphics[width=8.6cm]{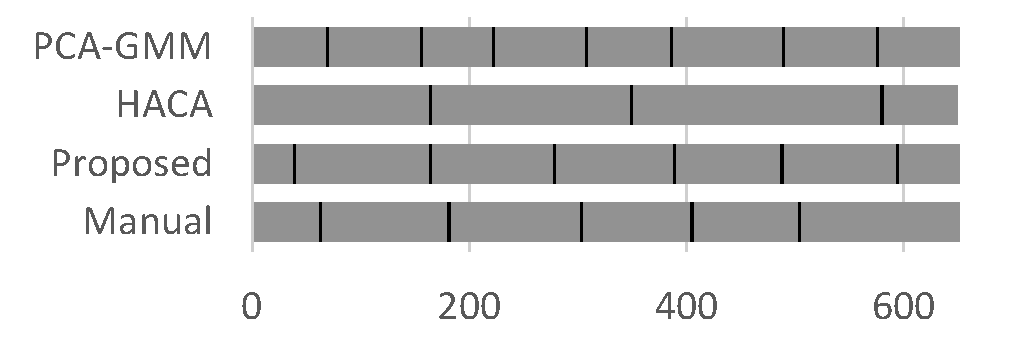}&
		\includegraphics[width=8.6cm]{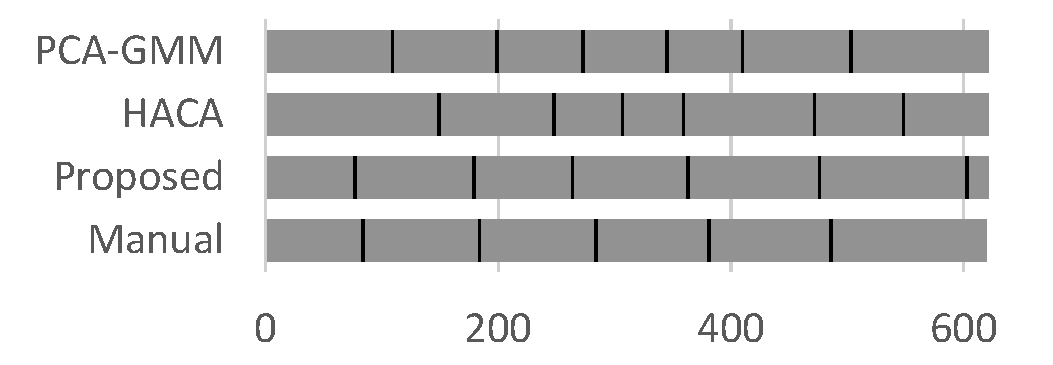}\\
		(a) action: ``\textit{Shallow Squats}'', actor: \#1 & (b) action: ``\textit{Shallow Squats}'', actor: \#8  \\
		\includegraphics[width=8.6cm]{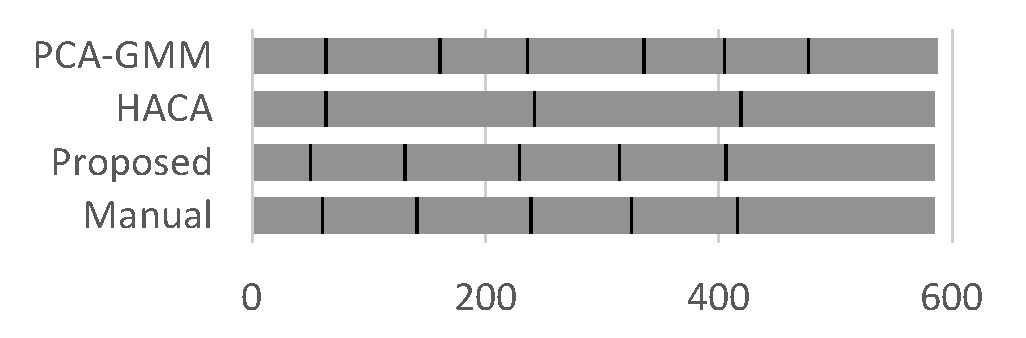}&
		\includegraphics[width=8.6cm]{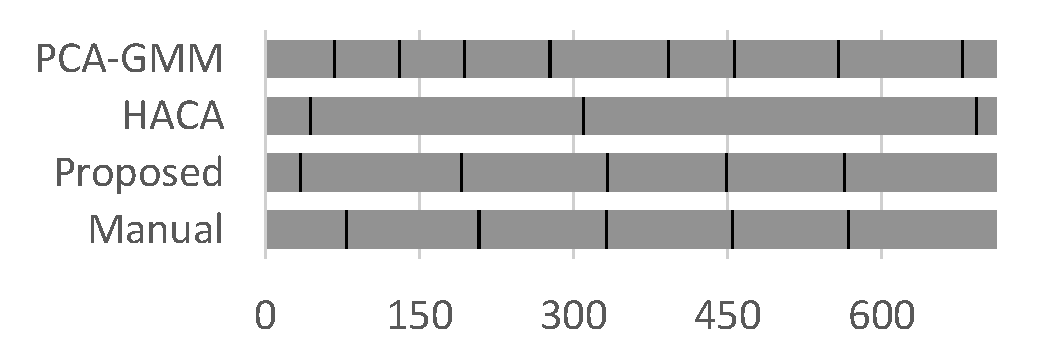}\\
		(c) action: ``\textit{Cops \& Robbers}'', actor: \#1  & (d) action: ``\textit{Cops \& Robbers}'', actor: \#8 \\
		\multicolumn{2}{c}{\includegraphics[width = 6cm]{fig10_legend.png}}\\
	\end{tabular}
	\caption{Results of the temporal repetition segmentation using four different segmentation methods, Manual, Proposed, HACA \cite{zhou2013hierarchical}, and PCA-GMM \cite{barbivc2004segmenting}, on the Kinect captured motion sequences from our database.}
	\label{fig8}
\end{figure*}

Figure \ref{fig8} demonstrate repetition segmentation results on the Kinect captured motion sequences in our database. Figures \ref{fig8}a and \ref{fig8}b show the results of the sequence ``\textit{Shallow Squats}'' which are performed by the two actors, \#1 and \#8, respectively. Figures \ref{fig8}c and \ref{fig8}d illustrate the results on the motion sequence ``\textit{Cops \& Robbers}'', also performed by the same two actors. For clarity, only the first 600 frames are shown in the figures. Based on the results on the noisy Kinect motion capture data, our proposed algorithm still obtains a robust repetition segmentation that approximately matches the results of the manual segmentation. Similar to the results with HDM05 dataset, the PCA-GMM algorithm still cannot distinguish the repetitions of the same action. Therefore, this method generates several false segmentation points with the Kinect motion sequences. Some under-segmentation appears in the results generated by HACA. Compared to the results on the motion capture data in HDM05 database, the performance of the HACA approach on the noisy Kinect motion data degrades significantly. Since HACA extracts the features based on the joint angles which are obtained by the raw input of the motion data, the noise in the raw motion data propagates to the calculated joint angles, resulting in the under-segmentation. Our proposed algorithm, on the other hand, reduces the noise by both UKF and band-pass filtering. Furthermore, only the most representative parameters are used for the segmentation point detection, making the algorithm more robust to noise in the input motion data. 

We further evaluate the segmentation accuracy of the three algorithms defined as follows:

\begin{equation}
\label{eq15}
\alpha=1/D \sum_{i=1}^{D}(1-e_i/L_i ).
\end{equation}
In (\ref{eq15}), $e_i$ denotes the absolute value of the difference between the $i$-th segment obtained by the selected segmentation algorithm and manual approach, $L_i$ denotes the length the $i$-th segment obtained by manual approach, $D$ denotes the minimal number of segments between the selected algorithm and the manual approach. In the case of $\alpha=1$, the length of detected segments by the algorithm would be the same as for the manual segmentation. Smaller $\alpha$ thus corresponds to larger segmentation errors.

%rotateArmsBothBackward3Reps
%elbowToKnee3RepsLelbowStart
\begin{table}\small
\begin{center}
\begin{tabular}{|l|c|c|c|}
\hline
Motion sequences & Proposed & HACA & PCA-GMM \\
\hline\hline
\textit{clapAboveHead5Reps} & \textbf{0.98} & \textbf{0.98} & 0.54 \\
\textit{clap5Reps} & \textbf{0.95} & 0.94 & 0.48 \\
\textit{jumpingJack3Reps} & 0.97 & \textbf{0.98} & 0.45 \\
\textit{rotateArmsBothBackward3Reps} & 0.96 & \textbf{0.97} &0.38 \\
\textit{elbowToKnee3RepsLelbowStart} & \textbf{0.92} & 0.91 & 0.36 \\
Average & \textbf{0.96} & \textbf{0.96} & 0.44 \\
\hline
\end{tabular}
\end{center}
\caption{Segmentation accuracy of the sequences from HDM05 database acquired by motion capture system}
\label{tb1}
\end{table}

\begin{table}
\begin{center}
\begin{tabular}{|l|c|c|c|}
\hline
Motion sequences & Proposed & HACA & PCA-GMM \\
\hline\hline
\textit{Shallow Squats} & \textbf{0.93} & 0.43 & 0.37 \\
\textit{Chair Stands} &\textbf{ 0.94} & 0.42 & 0.33 \\
\textit{Buddha's Prayer} & \textbf{0.91} & 0.43 & 0.35 \\
\textit{Cops \& Robbers} & \textbf{0.89} & 0.39 & 0.32 \\
\textit{Abs in Knee Lifts} & \textbf{0.92} & 0.40 & 0.39 \\
\textit{Lateral Stepping} & \textbf{0.87} & 0.38 & 0.35 \\
\textit{Clapping} & \textbf{0.90} & 0.44 & 0.41 \\
\textit{Punching} & \textbf{0.81} & 0.39 & 0.33 \\
\textit{Line Stepping} & \textbf{0.88} & 0.47 & 0.43 \\
\textit{Pendulum} & \textbf{0.85} & 0.46 & 0.40 \\
Average & \textbf{0.89} & 0.42 & 0.37 \\
\hline
\end{tabular}
\end{center}
\caption{Segmentation accuracy of the Kinect captured motion sequences from our database}
\label{tb2}
\end{table}

Table \ref{tb1} shows the segmentation accuracy of the three methods as compared to the manual approach on the motion sequences from the HDM05 database. Table \ref{tb2} shows the segmentation accuracy of the three methods for the Kinect captured motion sequences from our database. The values in Tables \ref{tb1} and \ref{tb2} represent the average across all the actors performing each sequence. For the motion data from motion capture database, our method achieves similar performance as HACA, and outperforms PCA-GMM. For the motion data in our database captured by Kinect, our approach still achieves the best performance among the three approaches. Since HACA also suffers in case of noisy data as PCA-GMM, the performance of HACA is similar as that of PCA-GMM, but significantly degraded compared to the performance of HACA on the motion capture data.

To demonstrate the impact of the noisy motion data from Kinect, we further evaluate the segmentation accuracy of the three methods with the same sequences as Table \ref{tb2} in our database but acquired by motion capture system. The results are summarized in Table \ref{tb3}. Compared to the performance on the Kinect data, the accuracy of our approach and PCA-GMM is slightly improved, while the accuracy of HACA is improved significantly. This result confirms our hypothesis that the noise in the Kinect capture data greatly affects the performance of HACA. However, the accuracy degradation of our method is much smaller compared to that of HACA.

\begin{table}
\begin{center}
\begin{tabular}{|l|c|c|c|}
\hline
Motion sequences & Proposed & HACA & PCA-GMM \\
\hline\hline
\textit{Shallow Squats} & \textbf{0.96} & \textbf{0.96} & 0.47 \\
\textit{Chair Stands} & 0.96 & \textbf{0.98} & 0.43 \\
\textit{Buddha’s Prayer} & 0.96 & \textbf{0.98} & 0.42 \\
\textit{Cops \& Robbers} & 0.94 & \textbf{0.97} & 0.40 \\
\textit{Abs in Knee Lifts} & 0.95 & \textbf{0.96} & 0.49 \\
\textit{Lateral Stepping} & 0.92 & \textbf{0.94} & 0.44 \\
\textit{Clapping} & \textbf{0.93} & \textbf{0.93} & 0.51 \\
\textit{Punching} & \textbf{0.90} & 0.88 & 0.41 \\
\textit{Line Stepping} & 0.93 & \textbf{0.97} & 0.53 \\
\textit{Pendulum} & \textbf{0.90} & 0.89 & 0.50 \\
Average & 0.94 & \textbf{0.95} & 0.46 \\
\hline
\end{tabular}
\end{center}
\caption{Segmentation accuracy of the sequences acquired by motion capture system from our database}
\label{tb3}
\end{table}

\begin{figure}[t]
	\begin{center}
		\includegraphics[width=9 cm]{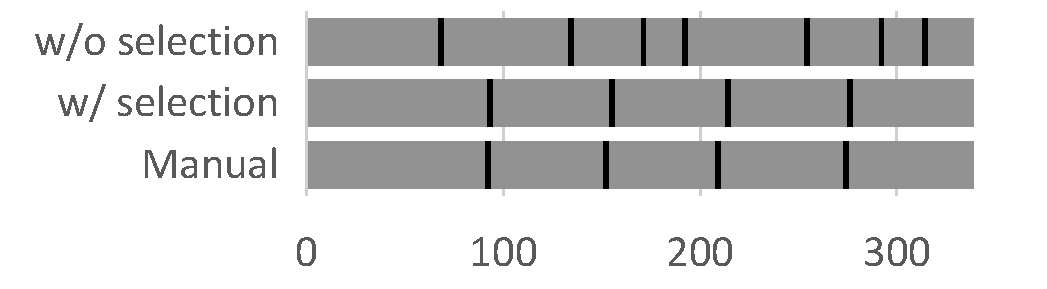} \\
		(a) action: ``\textit{clapOverHead5Reps}'', actor: ``bd'' \\
		\includegraphics[width=9 cm]{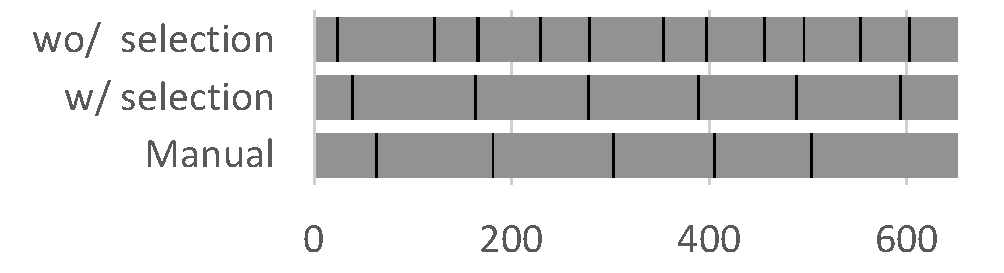} \\
		(b) action: ``\textit{Shallow Squats}'', actor: \#1 \\
	\end{center}
	\caption{Temporal repetition segmentation performance comparison between the proposed algorithm with and without most representative kinematic parameter selection.}
	\label{fig9}
\end{figure}

To verify the importance of the most representative kinematic parameter selection in our method, we perform the segmentation with and without the selection step. For the segmentation with kinematic parameter selection, the power selection threshold was set to 0.9. Figure \ref{fig9}a and \ref{fig9}b illustrate the segmentation results on the sequences ``\textit{clapOverHead5Reps}'' and ``\textit{Shallow Squats}'', respectively.

We observe that the proposed segmentation approach with parameter selection obtains quite similar results as the manual segmentation for both HDM05 and our databases. The segmentation without the parameter selection results in segmentation point shift and over-segmentation. The errors are caused by the interference from the parameters with inconsistent cyclic characteristics during the zero-velocity crossing detection.

\section{Conclusions}

In this paper, we proposed an unsupervised temporal repetition segmentation algorithm for human repetitive motion analysis that can be applied to various input modalities. The experimental results demonstrate that the proposed algorithm achieves robust repetition segmentation performance which is comparable to the labor-intensive manual segmentation on both the high precision motion capture data and the noisy motion data captured by the low-cost motion capture device like Kinect. Other state-of-the-art temporal action segmentation algorithms cannot distinguish the segments of repetitive action, like PCA-GMM, or suffer from the noise in the Kinect captured motion data, like HACA. Since the proposed method is generic and unsupervised, it can be widely applied to various motion capture modalities and various types of repetitive human activities.

{\small
\bibliographystyle{ieee}
\bibliography{egbib}
}

\end{document}